\documentclass{article}

\usepackage[nonatbib, final]{nips_2017}

\usepackage[utf8]{inputenc} 
\usepackage[T1]{fontenc}    
\usepackage{hyperref}       
\usepackage{url}            
\usepackage{booktabs}       
\usepackage{amsfonts}       
\usepackage{nicefrac}       
\usepackage{microtype}      
\usepackage{graphicx}
\usepackage[colorinlistoftodos]{todonotes}

\title{Sim-to-Real Transfer of Accurate Grasping with Eye-In-Hand Observations and Continuous Control}
\ifdefined\REVIEW

\author{
 Submission \#4, AIRW 2017}

\else

\author{
 M. Yan \\
 Department of Electrical Engineering \\
  Stanford University\\
  \texttt{mengyuan@stanford.edu}
  \And
  I. Frosio\\
  NVIDIA \\
  \texttt{ifrosio@nvidia.com}
  \And
  S. Tyree \\
  NVIDIA \\
  \texttt{styree@nvidia.com}
  \And
  J. Kautz \\
  NVIDIA \\
  \texttt{jkautz@nvidia.com}
}

\fi

\begin{document}

\maketitle

\begin{abstract}

In the context of deep learning for robotics, we show effective method of training a real robot to grasp a tiny sphere ($1.37$cm of diameter), with an original combination of system design choices.
We decompose the end-to-end system into a vision module and a closed-loop controller module. The two modules use target object segmentation as their common interface.
The vision module extracts information from the robot end-effector camera, in the form of a binary segmentation mask of the target.
We train it to achieve effective domain transfer by composing real background images with simulated images of the target.
The controller module takes as input the binary segmentation mask, and thus is agnostic to visual discrepancies between simulated and real environments.
We train our closed-loop controller in simulation using imitation learning and show it is robust with respect to discrepancies between the dynamic model of the simulated and real robot: when combined with eye-in-hand observations, we achieve a $90\%$ success rate in grasping a tiny sphere with a real robot.
The controller can generalize to unseen scenarios where the target is moving and even learns to recover from failures.

\end{abstract}

%
%
%
%
%
%
%
%
%


\section{Introduction}
\label{sec:Introduction}

Modern robots can be carefully scripted to execute repetitive tasks when 3D models of the objects and obstacles in the environment are known \emph{a priori}, but this same strategy fails to generalize to the more compelling case of a dynamic environment, populated by moving objects of unknown shapes and sizes.
One possibility to overcome this problem is given by a change of paradigm, with the adoption of Deep Learning (DL) for robotics. Instead of hard coding a sequence of actions, complex tasks can be learned through reinforcement~\cite{GuH16} or imitation learning~\cite{Jam17}.
But this also introduces new challenges that have to be solved to make DL for robotics effective.
Learning has to be safe, and cost- and time-effective.
It is therefore common practice to resort to robotic simulators for the generation of the training data.
Accurate simulations of kinematics, dynamics~\cite{Pen17}, and visual environment~\cite{Tob17} are required for effective training, but the computational effort grows with the simulation accuracy, slowing down the training procedure.
The overall learning time can be minimized by compromising between the accuracy of the simulator, the sample efficiency of the learning algorithm~\cite{Lev17}, and a proper balance of the computational resources of the system~\cite{Bab17}.
Finally, strategies learned in simulation have to generalize well to the real world, which justifies the research effort in the space of domain transfer~\cite{Tob17} for the development of general and reliable control policies. 

\begin{figure}
\centering
\includegraphics[width=0.95\linewidth]{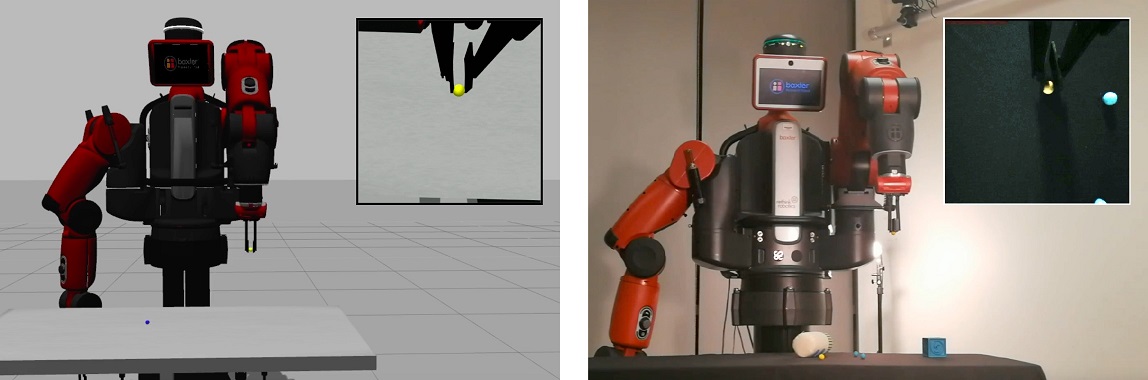}
\caption{We use imitation learning in simulation to train a closed-loop DNN controller, allowing a robot arm to successfully grasp a tiny (1.37cm of diameter) sphere (left panel).
The DNN controller's visual input is a binary segmentation mask of the target sphere, extracted from the RGB image captured by the end-effector camera (left inset).
A separate DNN vision module processes the real RGB images (right inset) to produce the same segmentation mask as in simulation, abstracting away the appearance differences between domains.
Combining the vision module and the controller module in a real environment (right panel), we achieve a grasping success rate of 90\% for the real robot.
}
\label{fig:teaser}
\end{figure}

This work focuses on grasping, a fundamental interaction between a robot and the environment.
We propose to decompose vision and control into separate network modules and use segmentation as their interface.
Our vision module generalizes across visual differences between simulated and real environments, and extracts the grasp target from the environment in the form of a segmentation mask.
The controller module, taking this domain-agnostic mask as input, is trained efficiently in simulation and applied directly in real environments.

Our system combines existing design choices from the literature in an original way. 
Our visual inputs are RGB images captured by the end-effector camera (eye-in-hand view), whereas most existing approaches use RGB or RGBD cameras observing the robot from a fixed, third-person point of view~\cite{Lev16, Ino17, Jam17, Tob17}.
We train a closed-loop DNN controller, whereas much of the literature use pre-planned trajectories to reach (and then grasp) an object~\cite{Mah16, Mah17, Pin16, Tob17}.
Finally, visual domain transfer is achieved by composing background images taken in the real environment with foreground objects from simulation and training the vision module to segment the target object from the composed images; our solution does not require rendering of complex 3D environments, as in~\cite{Jam17, Tob17}.

These choices are beneficial in several ways: the division into vision and control modules eases the interpretation of results, facilitates development and debugging, and potentially allows the re-use of the same vision or controller module for different robots or environments.
In fact, our DNN vision module is trained independently from the controller, using segmentation as an interface.
The combination of closed-loop control and eye-in-hand view allows us to successfully learn and execute a high precision task, even if the model of the simulated robot does not match the dynamics of the real robot. Using first person, eye-in-hand point of view allows effective refinement of state estimation as the robot arm approaches the target object, while the closed-loop controller compensates in real time for errors in the estimation of the sphere position and, at the same time, for errors in the dynamic response of the robot.
Achieving the same results with an open-loop controller, after observing the environment from a third person point of view, would be extremely challenging, as it would require an accurate estimate of the sphere position for trajectory planning, and the same accuracy during the execution of the trajectory.
Our real robot achieves $90\%$ success in grasping a $1.37$cm-diameter sphere, using RGB images and a closed-loop DNN controller.
The real robot is surprisingly robust to unseen clutter objects or a moving target, and has developed recovery strategies from failed grasp attempts.
This happens naturally without augmentation of the robot dynamics during training and with no LSTM module, which are otherwise required \cite{Pen17}.

\section{Related Work}
\label{sec:RelatedWork}


We give a brief review of several robotic systems aimed at grasping, each following a different approach to the system design. The effectiveness of the learning procedure is greatly affected by these design choices, including the selection of the sensors, the choice of an open- or closed-loop controller, the learning environment, and the learning algorithm. 

Different sensors provide measurements of the state of the robot and the environment at different levels of completeness and noise. While grasping with vision has a long history,
3D point clouds have been used as well, \emph{e.g.}\ in~\cite{Mah16,Mah17}. In the latter case, a CNN predicts the probability of grasp success from depth images.
Depth information is however not always available and sometimes difficult to extract from cluttered scenes or noisy environments.
A common configuration is to have an external RGB or RGBD camera observing the robot and the environment~\cite{Lev16, Ino17, Jam17, Tob17}; however, the camera pose relative to the robot has to be stable and carefully controlled to extract accurate geometric information and consequently guarantee the successful execution of the robot movements.

Open-loop controllers have been widely adopted in recent works, as they allow separation of the vision and control problems.
Such approaches require inspecting the environment once, \emph{e.g.}\ to detect the pose of the objects in the scene, and then using inverse kinematics to plan a trajectory to reach and grasp the desired object~\cite{Mah17, Pin16,Tob17}.
A closed-loop controller, on the other hand, allows recovery from failures and reactions to a dynamic environment. The price to be paid is the increased complexity of the controller, which generally takes as input both vision and kinematic data from the robot~\cite{Pen17}.

When it comes to the choice of the learning environment, robots can learn to grasp directly in the real world~\cite{Lev16, Sin17}, but training from scratch on real robots is costly, potentially unsafe, and requires very lengthy training times. 
Training in simulation is easier and cheaper, but transfer learning is needed to generalize from simulation to the real world.
The geometric configuration and appearance of the simulated environment can be extensively randomized, to create diverse training images that allow the neural network to generalize to previously unseen settings~\cite{Jam17}.
In~\cite{Ino17} a variational autoencoder changes the style of the real images into that of the corresponding simulated images; although effective, this method requires coupled simulated and real image pairs to learn the style transfer mapping, thus it does not scale to complex environments.
It is worth mentioning that the domain transfer problem does not apply only to vision: since the dynamics of a simulated robot may differ from its real-world counterpart, randomization can also be applied in this domain to facilitate generalization of the trained controller~\cite{Pen17, Tob17}.

Another design choice regards the learning algorithm.
The training speed depends on the cost of generating training data, the sample efficiency of the learning algorithm~\cite{Lev17}, and the balance of the available computational resources~\cite{Bab17}.
Deep RL algorithms have been successfully used to play Go and Atari games at superhuman level~\cite{Mnih15, Sil16, Sil17, Bab17}.
A3C is also employed for robotics in~\cite{Rus16}, to successfully train a DNN to control a robot arm that performs well in real environment. 
This requires the use of a progressive network~\cite{Rus16Progress}, using discrete actions instead of continuous control, and as many as $20$M frames of simulation data to reach convergence. 
Generally speaking, algorithms like A3C~\cite{Mnih15, Bab17} inefficiently explore the space of the possible solutions, slowing down the learning procedure if the cost of simulating the robot is high.
More sample efficient RL algorithms, like DDPG~\cite{Lil15}, explore the solution space more effectively and consequently move some of the computational demand from the simulation to the training procedure, but still require a huge amount of data to reach convergence.
The most sample efficient learning procedures are instead based on imitation learning: in this case the trained agent receives supervision from human demonstrations or from an oracle policy in simulation, thus the need for policy exploration is minimal and sample efficiency is maximized~\cite{Jam17, Sin17}.
Many imitation learning variants have been proposed to improve test-time performance and prevent exploding error~\cite{Ros10, Ros11}. We used DAGGER~\cite{Ros11} to train our DNN controller in simulation, with an expert designed as a finite state machine.

\section{Method}
\label{sec:Method}

\subsection{Imitation learning}
\label{subsec:dagger}

We use DAGGER ~\cite{Ros11}, an iterative algorithm for imitation learning, to learn a deterministic policy that allows a robot to grasp a $1.37$cm diameter yellow sphere.
Here we give a brief overview of DAGGER; for a detailed description we refer the readers to the original paper~\cite{Ros11}. 
Given an environment $E$ with state space $s$ and transition model $T(\mathbf{s}, \mathbf{a}) \rightarrow \mathbf{s}'$, we want to find a policy $\mathbf{a} = \pi(\mathbf{s})$ that reacts to every observed state $\mathbf{s}$ 
in the same manner as an expert policy $\pi_E$. 
During the first iteration of DAGGER, we gather a dataset of state-action pairs by executing the expert policy $\pi_E$ and use supervised learning to train a policy $\pi_1$ to reproduce the expert actions.
At iteration $n$, the learned policy $\pi_{n-1}$ is used to interact with the environment and gather observations, while the expert is queried for the optimal actions on the states observed and new state-action pairs are added to the dataset.
The policy $\pi_n$ is initialized from $\pi_{n-1}$ and trained to predict expert actions on the entire dataset.
At each iteration the gathered state observation distribution is induced by the evaluated policy, and over time the training data distribution will converge to the induced state distribution of the final trained policy. 

\subsection{Design of the DAGGER expert}
\label{subsec:expert}

We train our DNN controller in simulation, using Gazebo~\cite{Koe04} to simulate a Baxter robot.
The robot arm is controlled in position mode; we indicate with $[s_0, s_1, e_0, e_1, w_0, w_1, w_2]$ the seven joint angles of the robot arm and with $[g_0]$ the binary command to open or close the gripper.
Controller learning is supervised by an expert which, at each time step, observes joint angles and gripper state, as well as the simulated position of the target sphere.
The expert implements a simple but effective finite-state machine policy to grasp the sphere.
In state $s_0$, the end-effector moves along a linear path to a point $6$cm above the sphere; in state $s_1$, the end-effector moves downward to the sphere; when the sphere center is within $0.1$cm from the gripper center, the expert enters into state $s_2$ and closes the gripper. 
In case the gripper accidentally hits the sphere or fails to grasp it (this can happen because of simulation noise, inaccuracies, or grasping a non-perfectly centered sphere), the expert goes back to $s_0$ or $s_1$ depending on the new sphere position. A video of the expert is shown at \url{https://youtu.be/5JffwreA47o}.


\subsection{Controller DNN architecture and training}
\label{subsec:training}

\begin{figure}
\begin{center}
\includegraphics[width=0.9\textwidth]{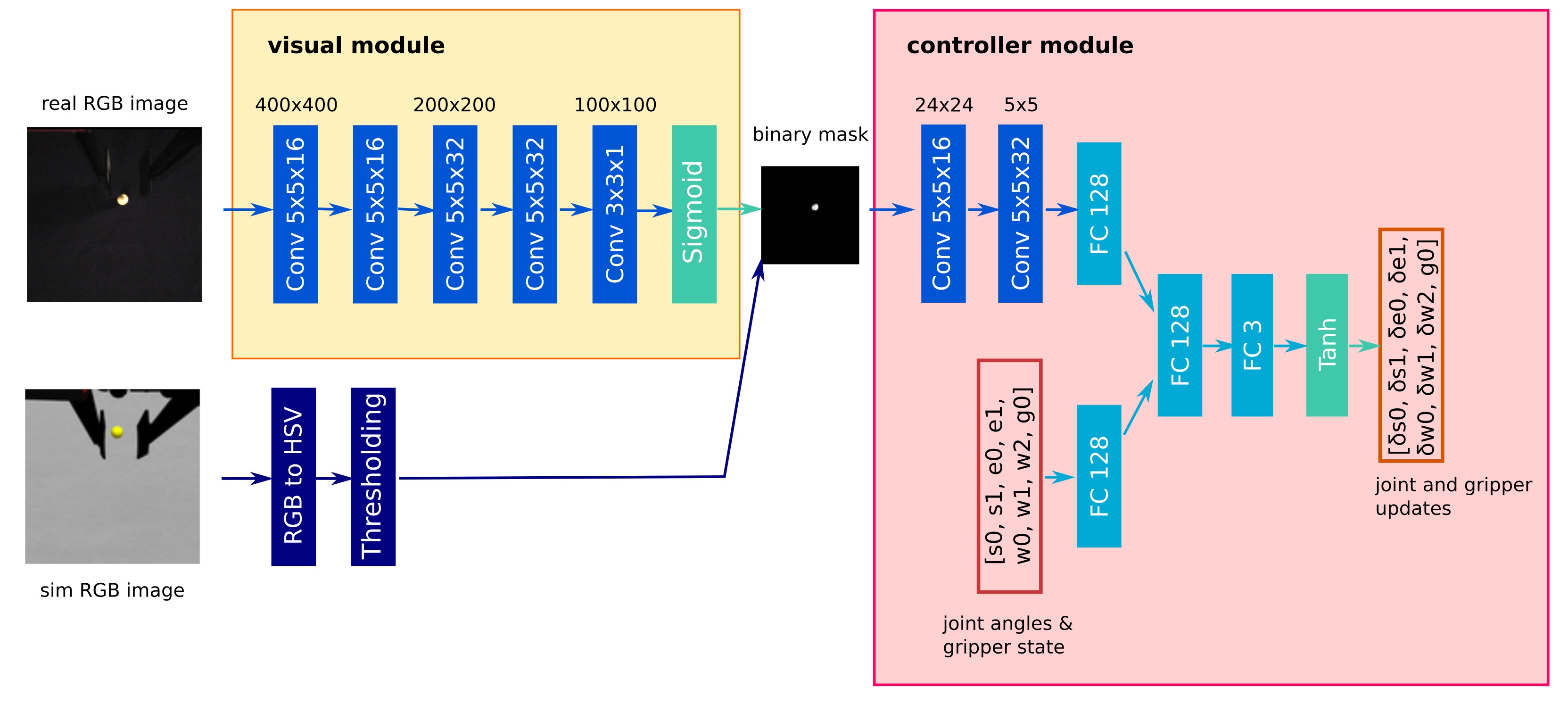}
\caption{The DNNs for the vision module (left) and closed-loop controller module (right). 
The vision module takes RGB images from the end-effector camera and label the sphere pixels. The control module  takes as input the segmentation mask and the current configuration of the robot arm (7 joint angles plus the gripper status), and it outputs an update for the robot configuration. 
}
\label{fig:network}
\end{center}
\end{figure}

Our closed-loop DNN controller processes the input information along two different pathways, one for the visual data and the other for the robot state.
The visual input is a $100 \times 100$ segmentation mask, obtained by cropping and downsampling the RGB image captured by the end-effector camera and segmenting the target object from background via one of the methods described in Section \ref{subsec:transfer}.
The resulting field of view is approximately $80$ degrees.
The segmentation mask is processed by $2$ convolutional layers with $16$ and $32$ filters respectively, each of size $5 \times 5$ and with stride $4$, and a fully connected layer with $128$ elements; ReLU activations are applied after each layer (see Fig. \ref{fig:network}).
The robot state pathway has one fully connected layer (followed by ReLU) that expands the $8$ dimensional vector of joint angles and gripper state into a $128$ dimensional vector.
The outputs of the two pathways are concatenated and fed to $2$ additional fully connected layers to output the action command, \emph{i.e.} the changes of joint angles and gripper status $[\delta s_0, \delta s_1, \delta e_0, \delta e_1, \delta w_0, \delta w_1, \delta w_2, g_0]$.
A \emph{tanh} activation function limits the absolute value of each command.
Contrary to~\cite{Jam17}, we do not use an LSTM module and do not observe a negative impact on the final result, although a memory module could help recovering when the sphere is occluded from the view of the end-effector camera.

To train this DNN with DAGGER, we collect $1000$ frames at each iteration, roughly corresponding to $10$ grasping attempts.
At each iteration we run $200$ epochs of ADAM on the accumulated dataset, with a learning rate of $0.001$ and batch size $64$.
The training loss in DAGGER is the squared L2 norm of the difference between the output of the DNN and the ground truth actions provided by the expert agent, thus defined at iteration $n$ as:
\begin{equation}
\mathcal{L} = ||\pi_n(\mathbf{s}) - \pi_E(\mathbf{s})||^2
\label{eq:L}
\end{equation}
%
Partially inspired by~\cite{Jam17}, we also tested an augmented cost function including two auxiliary tasks, forcing the network to predict the sphere position in the end-effector coordinate frame from the input segmentation mask, and the position of the end-effector in the world reference frame from the robot state.
However, we found that these auxiliary tasks did not help the network convergence nor the rate of successful grasps, so we decided not to include any of them in the cost function.

%

\subsection{Vision module and domain transfer}
\label{subsec:transfer}


To ensure that the geometric information captured by the simulated and real cameras are coherent with each other, we calibrate the internal parameters (focal length, principal points) of the end-effector camera on the real robot and apply the same parameters in the Gazebo simulation.
We do not apply any distortion correction as we assume that a good policy executed with a closed-loop control tends to see the sphere in the center of the image, where distortions are minimal.
Beyond geometry, image appearance plays a fundamental role in the domain transfer problem.
As shown in insets of Fig. \ref{fig:teaser}, the Gazebo images are noise-free and well-lit, contain objects with sharp color and no textures, as well as poorly rendered shadows, especially when the end-effector is close to the sphere.
On the other hand, the images taken by the Baxter camera are often dark, noisy, and blurry, with the robot gripper casting heavy and multiple shadows on the sphere and table.

We test two methods to enable generalization from simulated to real robots under these visual differences.
In our baseline method, we work in the HSV color space and apply a manually set threshold to the H and S channels, to extract the sphere pixels, and pass the segmentation mask to the DNN controller (Fig. \ref{fig:network}), previously trained with DAGGER on the same input.
This method is quick to implement and validate, and it achieves perfect segmentation of the sphere on the simulated images.
Generalization to the real environment is achieved by re-tuning the threshold, and in our controlled experiments can achieve almost perfect results on the real images.
This helps analyzing the effects of dynamic and visual domain differences separately.
Nonetheless, the threshold mechanism does not work well if the environment conditions change, and cannot be easily applicable to more complex target objects.
For instance, when a human hand is present in the camera field-of-view, part of the hand is recognized as sphere and the DNN controller consequently fails in grasping the sphere.

Our second method is inspired by \emph{domain randomization}~\cite{Jam17, Tob17}.
We use a DNN vision module composed of five convolutional layers (Fig. \ref{fig:network}) to process the $400 \times 400$ RGB input images and generate $100 \times 100$ binary segmentation masks.
Training data are generated by alpha-blending simulated images of the sphere in random positions with image backgrounds taken from the real robot end-effector camera; blending is performed based on the simulated sphere's segmentation mask.
We collect $800$ images of the real environment by setting the robot's arm at random poses in the robot's workspace.
The collection procedure is completely automated and takes less than $30$ minutes.
During training, the combined background and sphere images are randomly shifted in the HSV space to account for the differences in lighting, camera gain, and color balance. 
Cross-entropy loss is used as a cost function, and early stopping is used to prevent overfitting, using a small set of real images as validation.
Our randomization procedure is different from the one used in~\cite{Tob17, Jam17}: since our method only requires blending random background images with the simulated sphere, it greatly reduces the time to design a simulator to generate sufficiently diverse training data.

\begin{figure}
\begin{center}
\includegraphics[width=\textwidth]{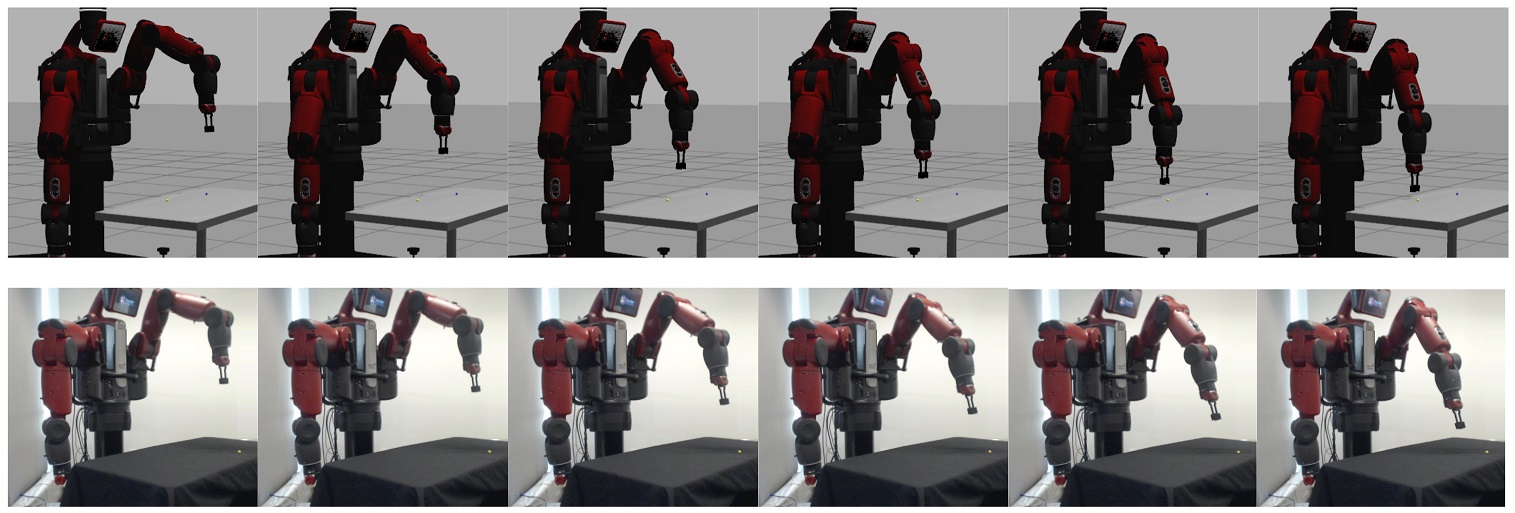}
\caption{Starting from the same initial configuration (left-most panels), the simulated (top row) and real (bottom row) robots execute the same sequence of $150$ commands at $10$Hz. Because of differences in the dynamic model of the simulated and real robot, the final configurations of the two robots are different. Because of this, we adopt a closed-loop controller to allow for error compensation while executing the trajectory, dramatically reducing the negative effects of this ``reality gap.''}
\label{fig:control_issues}
\end{center}
\end{figure}

A second aspect potentially requiring domain adaptation is the ``reality gap'' between the dynamic response of the simulated and real robots.
Several issues may contribute to the generation of this gap, including an inaccurate robot model or model parameter setting, hysteresis, joint friction (hard to calibrate or model), delays in the transmission of the control signals, and noisy measurements of the state in the real robot~\cite{Tob17}. 
Fig. \ref{fig:control_issues} shows how the responses of the real and simulated robots can differ with an open-loop controller: starting from the same configuration, the execution of the same sequence of commands at the same frequency leads to two different robot configurations.
While small differences accumulate over time in the case of an open-loop controller, our choice of a closed-loop controller corrects execution errors online, leading to a stable and accurate system as shown in Section~\ref{sec:Results}, without requiring any dynamic domain adaptation as in~\cite{Pen17}.


\section{Results and Discussion}
\label{sec:Results}

\subsection{Grasping in simulation}
\begin{figure}
\begin{center}
\includegraphics[width=0.7\textwidth]{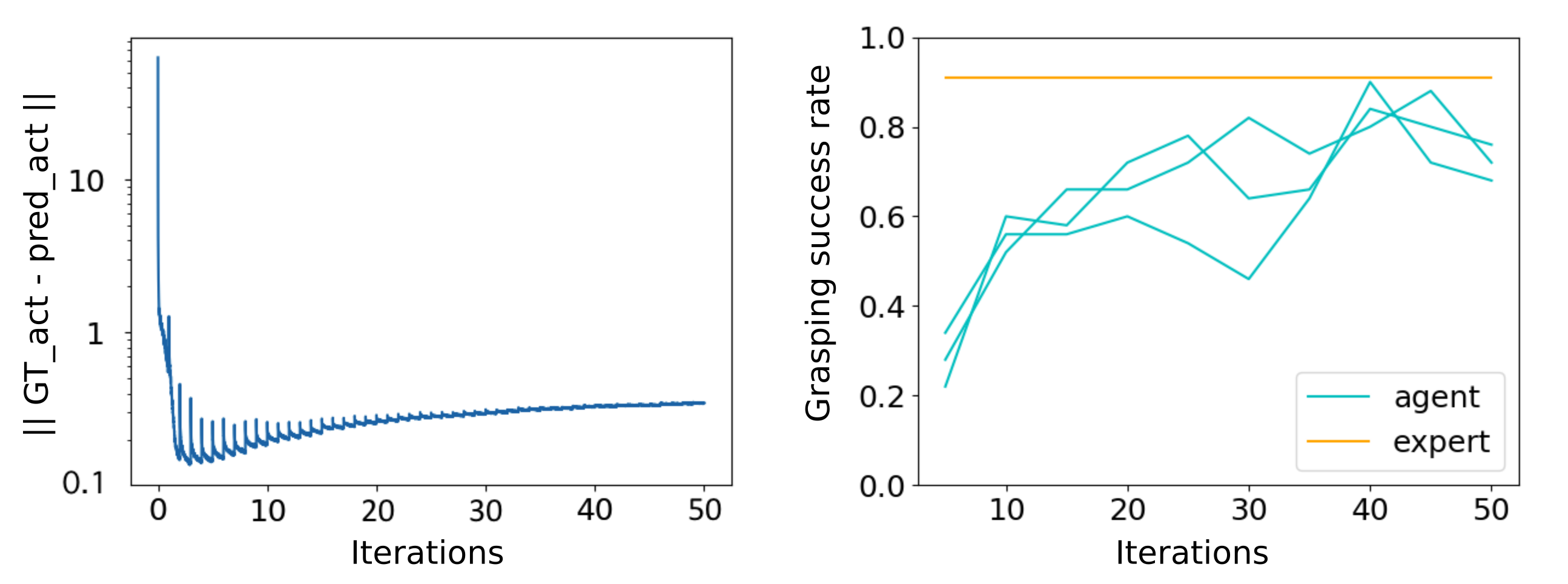}
\caption{The left panel shows the DAGGER cost function $\mathcal{L}$ in Eq. (\ref{eq:L}) during training. Overfitting occurs in early iterations when the dataset is small; then the cost stabilizes around its optimal value. The right panel shows the grasping success rate of the DNN controller in simulation, evaluated three times every five iterations of DAGGER. Each iteration adds $1000$ frames to the dataset.}
\label{fig:simEval}
\end{center}
\end{figure}

We evaluate our DNN controller module in simulation at the end of each iteration of DAGGER.
Evaluation is performed by measuring the number of successful grasps for a sphere located at $50$ positions regularly spaced on a rectangular grid.
For each position, the trial ends with a successful grasp or after $150$ steps.
To account for uncertainties in the simulator, we run the evaluation three times.
Fig. \ref{fig:simEval} shows the the grasping success rate as training progresses: $50$K training frames are sufficient to achieve a $90\%$ success rate, matching the performance of the expert.
Compared to~\cite{Rus16, Pop17, Jam17} that take $0.3$M, $50$M, and $1$M frames respectively to solve similar tasks, we see the superior data efficiency of DAGGER, relative to other reinforcement or imitation learning algorithms.

Visual inspection of failed attempts reveals that on rare occasions the grippers collide with the table and cannot be closed; such cases could be solved if force sensors are available on the robot end-effector.
However, in most failure cases, the robot's end-effector reaches the sphere but touches it causing the sphere to roll away and out of the camera field of view, too far to be reached even from the expert.
Reinforcement learning algorithms have the potential to explore policies for such corner cases, or a more expressively scripted expert could offer guidance.
Reinforcement learning from scratch requires a much longer training time to effectively explore the space of possible solutions~\cite{Lev17}, while programming time is needed to design a more complex expert.
A combination of both methods may have merit:
imitation learning with a simple expert can learn an effective control policy in a short amount of time, while reinforcement learning may be engaged as a second step to solve for corner cases and improve the robustness of the learned policy.
A video of our closed-loop DNN controller in action in simulation can be seen at \url{https://youtu.be/liF4FktsxOQ}.

\begin{figure}
\begin{center}
\includegraphics[width=0.8\textwidth]{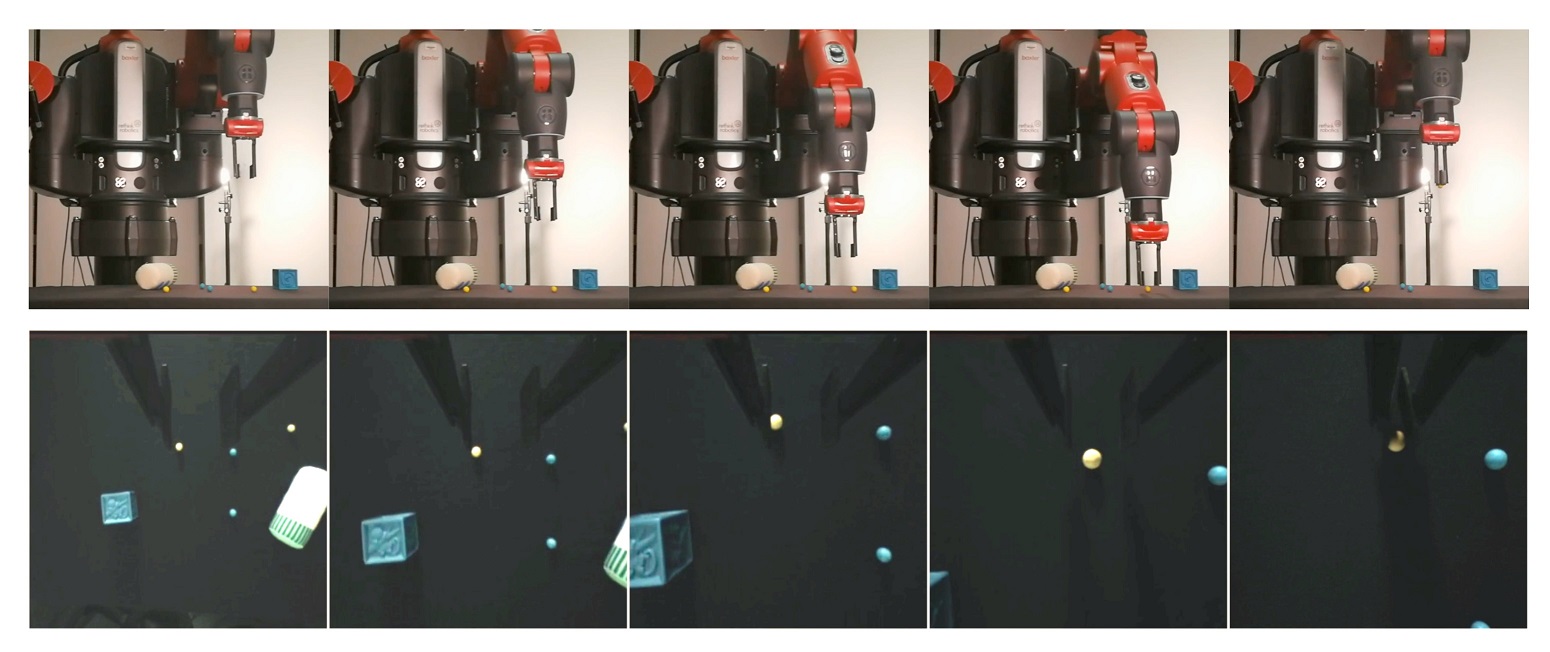}
\caption{Snapshots of the learned agent grasping in a real environment. The only visual input of the DNN closed-loop controller is the end-effector camera image shown in the bottom row. We have modified the brightness and contrast of the images for ease of viewing.}
\label{fig:realResult}
\end{center}
\end{figure}

\subsection{Grasping with a real robot}

The DNN controller trained in simulation and tested directly on the real robot without further training or fine-tuning is surprisingly robust to differences in dynamics: using our baseline segmentation method to extract the segmentation masks of the sphere in a controlled environment, the real robot achieves an $80\%$ success rate on $20$ grasping attempts with the sphere in random positions. 
This result is achieved thanks to the closed-loop approach we take for the controller: since the controller corrects previous position errors, the ``reality gap'' between the dynamics of the simulated and real robots does not represent a critical issue, at least for a robot moving at limited speed. 

When testing the DNN controller on the real robot, we also observe the emergence of recovery strategies from failed attempts, when the controller raises the end-effector slightly above the table to relocate the target sphere.
Such recovery behaviour can only be scripted in the case of an open-loop formulation, as shown in~\cite{Lev16}, but it is learned automatically as an effect of the combined choices of a closed-loop controller, learning through DAGGER, and the design of the expert as a finite state machine. 
In fact, imitation learning using only expert demonstrations may fail to capture such rare behavior, since the expert mostly succeeds at the first attempt and thus only rarely include recovery behaviour.
On the other hand, when training with DAGGER, at iteration $n$ the DNN controller executes the sub-optimal policy $\pi_n$ to collect new training data, so it can expose and learn to correct its own errors by querying the expert agent for advice.
As a result, the training data covers a larger state distribution than the distribution induced by expert demonstrations, and the trained agent effectively learns to recover from possible failures.
A video of the robot acting in the real environment and showing such behavior can be seen in \url{https://youtu.be/mMUwz0zpb0Y}.

\subsection{Generalizing to visual domain differences}
\begin{figure}
\begin{center}
\includegraphics[width=0.8\textwidth]{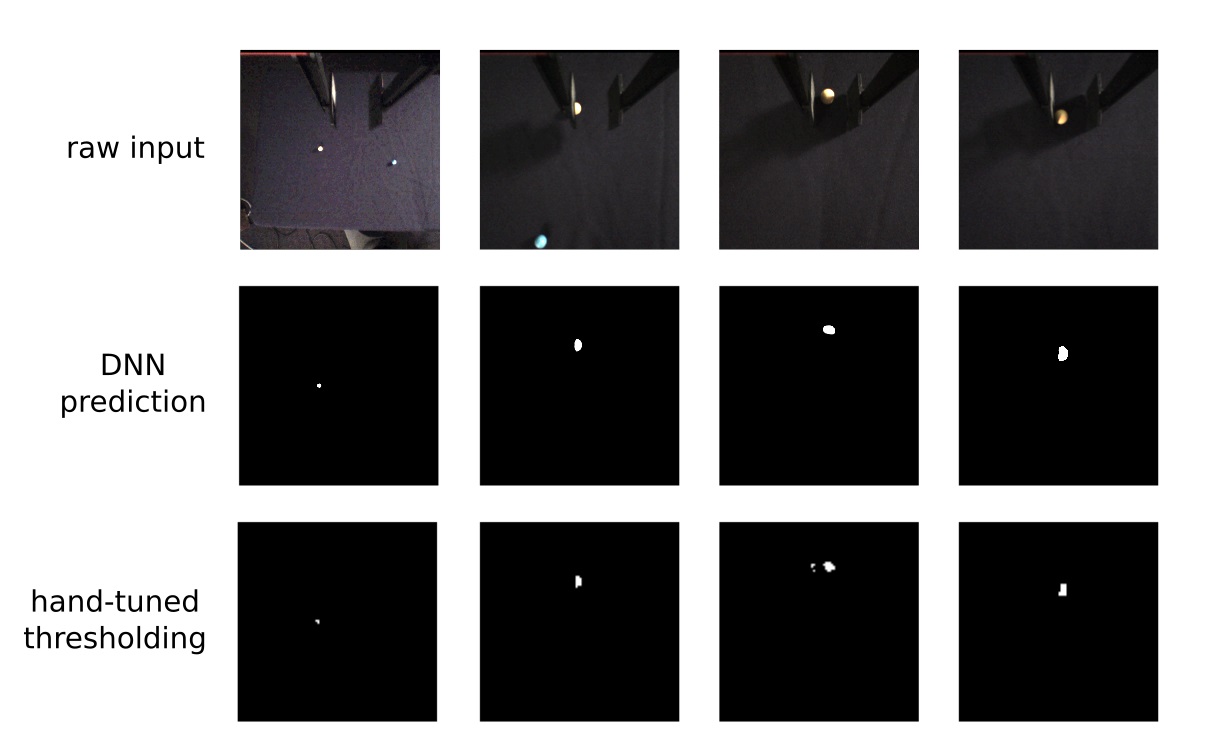}
\caption{The top row shows RGB images from the real robot end-effector camera (enhanced for visualization). The middle row shows segmentation masks for the yellow sphere, generated by hand-tuned threshold in the HSV color space (baseline method). The bottom row shows the output segmentation of our DNN vision module. The DNN correctly identifies the sphere pixels when it is partially occluded (second column), discriminates the target sphere from the yellowish gripper (third column), and recognizes more pixels in case of shadows (fourth column).}
\label{fig:segmentation_eval}
\end{center}
\end{figure}

An agent that is more robust to changes in the environment, \emph{e.g.} lighting conditions, is obtained with the introduction of the DNN vision module in Fig.~\ref{fig:network} trained with the proposed domain transfer technique.
We first evaluate the DNN vision module on a set of $2140$ images from the real robot, collected by running the DNN controller while using the segmentation masks from our baseline method.
We compare methods by adopting the masks from our baseline color filter as ground truth and comparing them to the output of the DNN vision module, though the baseline misidentifies some sphere pixels as background when the ball is in heavy shadow cast by the gripper.
When the output sigmoid layer of the segmentation network is thresholded at $0.5$, the DNN vision module achieves $85.3\%$ precision and $98.3\%$ recall compared to the baseline.
Visual comparisons (Fig. \ref{fig:segmentation_eval}) show the DNN identifies more sphere pixels, especially when the sphere is shadowed, and discriminates the illuminated gripper, which may also appear yellow, possibly using shape cues.

When the DNN controller uses the output of the DNN vision module instead of the baseline segmentation, the real robot can successfully grasp the sphere $90\%$ of the time.
More interestingly, the robot can reach the sphere even  when it moves (thanks to the closed-loop controller), or 
if multiple spheres or other clutter are in the field of view. 
Clutter objects never appear in the training set of the DNN vision module, which nevertheless differentiates between the yellow target sphere and other yellowish objects, \emph{e.g.} a hand, through better color or shape discrimination. 
In the same situation, our baseline method generates a segmentation mask with many false positives and the DNN controller fails to grasp the sphere. 
Snapshots from one successful grasp are shown in Figure~\ref{fig:realResult}, while a video can be seen in \url{https://youtu.be/mMUwz0zpb0Y}.

\section{Conclusion}
\label{sec:Discussion}

We present a method for training a robot to grasp a tiny sphere in simulation and transferring the learned controller to real environments.
We decompose the system into a vision module and a closed-loop controller module.
The vision module translates real RGB in-hand images into a segmentation of the target. 
The controller takes the segmentation mask, responding to the changing environment and robot state in a closed-loop manner, automatically adjusting to differences in dynamics between simulated and real robots. 
This modular design makes the system more interpretable 
and supports easier adaptation to new robots or visual environments, since only part of the system needs to be retrained.
We demonstrate efficient training of the vision module by composing simulated and real images, minimizing the data collection effort. 
The resulting system achieves a $90\%$ success rate in grasping a tiny sphere when tested on the real robot.
The system is robust to moving targets and background clutter and is often able to recover from failed grasp attempts.

In the future we plan to investigate how the binary segmentation can be generalized to multi-label segmentation, with applications to robotic tasks where multiple objects and their relations need to be considered, e.g., stacking cubes. We also plan to generalize our domain transfer method for objects with different shapes and colors, compare our modular approach with end-to-end learning, and consider the application of reinforcement learning for the fine-tuning of the learned policy.

\bibliography{references}

\begin{thebibliography}{10}

\bibitem{Bab17}
{\sc Babaeizadeh, M., Frosio, I., Tyree, S., Clemons, J., and Kautz, J.}
\newblock Reinforcement learning thorugh asynchronous advantage actor-critic on
  a gpu.
\newblock In {\em ICLR\/} (2017).

\bibitem{GuH16}
{\sc Gu, S., Holly, E., Lillicrap, T.~P., and Levine, S.}
\newblock Deep reinforcement learning for robotic manipulation.
\newblock {\em CoRR abs/1610.00633\/} (2016).

\bibitem{Ino17}
{\sc Inoue, T., Chaudhury, S., De~Magistris, G., and Dasgupta, S.}
\newblock Transfer learning from synthetic to real images using variational
  autoencoders for robotic applications.
\newblock {\em arXiv preprint arXiv:1709.06762\/} (2017).

\bibitem{Jam17}
{\sc James, S., Davison, A.~J., and Johns, E.}
\newblock Transferring end-to-end visuomotor control from simulation to real
  world for a multi-stage task.
\newblock {\em CoRR abs/1707.02267\/} (2017).

\bibitem{Koe04}
{\sc Koenig, N., and Howard, A.}
\newblock Design and use paradigms for gazebo, an open-source multi-robot
  simulator.
\newblock In {\em IEEE/RSJ International Conference on Intelligent Robots and
  Systems\/} (Sendai, Japan, Sep 2004), pp.~2149--2154.

\bibitem{Lev17}
{\sc Levine, S., and Finn, C.}
\newblock Deep reinforcement learning, decision making, and control.
\newblock 2017.

\bibitem{Lev16}
{\sc Levine, S., Pastor, P., Krizhevsky, A., and Quillen, D.}
\newblock Learning hand-eye coordination for robotic grasping with deep
  learning and large-scale data collection.
\newblock {\em CoRR abs/1603.02199\/} (2016).

\bibitem{Lil15}
{\sc Lillicrap, T.~P., Hunt, J.~J., Pritzel, A., Heess, N., Erez, T., Tassa,
  Y., Silver, D., and Wierstra, D.}
\newblock Continuous control with deep reinforcement learning.
\newblock {\em CoRR abs/1509.02971\/} (2015).

\bibitem{Mah17}
{\sc Mahler, J., Liang, J., Niyaz, S., Laskey, M., Doan, R., Liu, X., Ojea,
  J.~A., and Goldberg, K.}
\newblock Dex-net 2.0: Deep learning to plan robust grasps with synthetic point
  clouds and analytic grasp metrics.
\newblock {\em arXiv preprint arXiv:1703.09312\/} (2017).

\bibitem{Mah16}
{\sc Mahler, J., Pokorny, F.~T., Hou, B., Roderick, M., Laskey, M., Aubry, M.,
  Kohlhoff, K., Kr{\"o}ger, T., Kuffner, J., and Goldberg, K.}
\newblock Dex-net 1.0: A cloud-based network of 3d objects for robust grasp
  planning using a multi-armed bandit model with correlated rewards.
\newblock In {\em Robotics and Automation (ICRA), 2016 IEEE International
  Conference on\/} (2016), IEEE, pp.~1957--1964.

\bibitem{Mnih15}
{\sc Mnih, V., Kavukcuoglu, K., Silver, D., Rusu, A.~A., Veness, J., Bellemare,
  M.~G., Graves, A., Riedmiller, M., Fidjeland, A.~K., Ostrovski, G., et~al.}
\newblock Human-level control through deep reinforcement learning.
\newblock {\em Nature 518}, 7540 (2015), 529--533.

\bibitem{Pen17}
{\sc Peng, X.~B., Andrychowicz, M., Zaremba, W., and Abbeel, P.}
\newblock Sim-to-real transfer of robotic control with dynamics randomization.
\newblock {\em arXiv preprint arXiv:1710.06537\/} (2017).

\bibitem{Pin16}
{\sc Pinto, L., and Gupta, A.}
\newblock Supersizing self-supervision: Learning to grasp from 50k tries and
  700 robot hours.
\newblock In {\em Robotics and Automation (ICRA), 2016 IEEE International
  Conference on\/} (2016), IEEE, pp.~3406--3413.

\bibitem{Pop17}
{\sc Popov, I., Heess, N., Lillicrap, T., Hafner, R., Barth-Maron, G., Vecerik,
  M., Lampe, T., Tassa, Y., Erez, T., and Riedmiller, M.}
\newblock Data-efficient deep reinforcement learning for dexterous
  manipulation.
\newblock {\em arXiv preprint arXiv:1704.03073\/} (2017).

\bibitem{Ros10}
{\sc Ross, S., and Bagnell, D.}
\newblock Efficient reductions for imitation learning.
\newblock In {\em Proceedings of the thirteenth international conference on
  artificial intelligence and statistics\/} (2010), pp.~661--668.

\bibitem{Ros11}
{\sc Ross, S., Gordon, G.~J., and Bagnell, D.}
\newblock A reduction of imitation learning and structured prediction to
  no-regret online learning.
\newblock In {\em International Conference on Artificial Intelligence and
  Statistics\/} (2011), pp.~627--635.

\bibitem{Rus16Progress}
{\sc Rusu, A.~A., Rabinowitz, C.~N., Desjardins, G., Soyer, H., Kirkpatrick,
  J., Kavukcuoglu, K., Pascanu, R., and Hadsell, R.}
\newblock {Progressive Neural Networks}.
\newblock {\em ArXiv e-prints\/} (June 2016).

\bibitem{Rus16}
{\sc Rusu, A.~A., Vecerik, M., Roth{\"o}rl, T., Heess, N., Pascanu, R., and
  Hadsell, R.}
\newblock Sim-to-real robot learning from pixels with progressive nets.
\newblock {\em arXiv preprint arXiv:1610.04286\/} (2016).

\bibitem{Sax08}
{\sc Saxena, A., Driemeyer, J., and Ng, A.~Y.}
\newblock Robotic grasping of novel objects using vision.
\newblock {\em The International Journal of Robotics Research 27}, 2 (2008),
  157--173.

\bibitem{Sil16}
{\sc Silver, D., Huang, A., Maddison, C.~J., Guez, A., Sifre, L., Van
  Den~Driessche, G., Schrittwieser, J., Antonoglou, I., Panneershelvam, V.,
  Lanctot, M., et~al.}
\newblock Mastering the game of go with deep neural networks and tree search.
\newblock {\em Nature 529}, 7587 (2016), 484--489.

\bibitem{Sil17}
{\sc Silver, D., Schrittwieser, J., Simonyan, K., Antonoglou, I., Huang, A.,
  Guez, A., Hubert, T., Baker, L., Lai, M., Bolton, A., et~al.}
\newblock Mastering the game of go without human knowledge.
\newblock {\em Nature 550}, 7676 (2017), 354--359.

\bibitem{Sin17}
{\sc Singh, A., Yang, L., and Levine, S.}
\newblock Gplac: Generalizing vision-based robotic skills using weakly labeled
  images.
\newblock {\em arXiv preprint arXiv:1708.02313\/} (2017).

\bibitem{Tob17}
{\sc Tobin, J., Fong, R., Ray, A., Schneider, J., Zaremba, W., and Abbeel, P.}
\newblock Domain randomization for transferring deep neural networks from
  simulation to the real world.
\newblock {\em CoRR abs/1703.06907\/} (2017).

\end{thebibliography}
\bibliographystyle{acm}

\end{document}